\documentclass[11pt,a4paper]{article}
\usepackage[hyperref]{acl2021}
\usepackage{times}
\usepackage{latexsym}

\usepackage{graphicx}
\usepackage{multirow}
\usepackage{booktabs}
\usepackage{hyperref}
\usepackage{float}
\usepackage{bbding}
\usepackage{pifont}
\usepackage{wasysym}
\usepackage{amssymb}

\usepackage{microtype}
\global\hbadness=10000

\makeatletter
\newcommand*\bigcdot{\mathpalette\bigcdot@{.8}}
\newcommand*\bigcdot@[2]{\mathbin{\vcenter{\hbox{\scalebox{#2}{$\m@th#1\bullet$}}}}}
\makeatother

\aclfinalcopy 
\def\aclpaperid{3404} 

\title{Grounding \textit{`Grounding'} in NLP}

\author{
Khyathi Raghavi Chandu,
Yonatan Bisk, 
Alan W Black \\
Language Technologies Institute \\
Carnegie Mellon University \\
\texttt{\{kchandu, ybisk, awb\}@cs.cmu.edu}
}

\date{}

\begin{document}
\maketitle
\begin{abstract}


The NLP community has seen substantial recent interest in grounding to 
facilitate interaction between language technologies and 
the world. However, as a community, we use the term broadly to reference \textit{any} linking of text to data or non-textual modality. In contrast, Cognitive Science more formally defines ``grounding" as \textbf{the process of establishing what mutual information is required for successful communication between two interlocutors} -- a definition which might implicitly capture the NLP usage but differs in intent and scope. 

We investigate the gap between these definitions and seek answers to the following questions: \textit{(1) What aspects of grounding are missing from NLP tasks?} 
Here we present the dimensions of coordination, purviews and constraints. 
\textit{(2) How is the term ``grounding'' used in the current research?} 
We study the trends in datasets, domains, and tasks introduced in recent NLP conferences.
And finally, \textit{(3) How to advance our current definition to bridge the gap with Cognitive Science?} We present ways to both create new tasks or repurpose existing ones to make advancements towards achieving a more complete sense of grounding.
{\small \href{https://github.com/khyathiraghavi/Grounding-Grounding}{github.com/khyathiraghavi/Grounding-Grounding} }


\end{abstract}

\section{Introduction}

We as humans communicate and interact for a variety of reasons with a goal. We use language to seek and share information, clarify misunderstandings that conflict with our prior knowledge and contextualize based on the medium of interaction to develop and maintain social relationships. 
However, language is only one of the enablers of this communication reliant on several auxiliary signals and sources such as documents, media, physical context etc., This linking of concepts to context is \textit{grounding} and within NLP context is often a knowledge base, images or discourse. 


\begin{figure}[t!]
\centering
\includegraphics[trim=1.2cm 2.2cm 11cm 1.5cm, clip,width=0.85\linewidth]{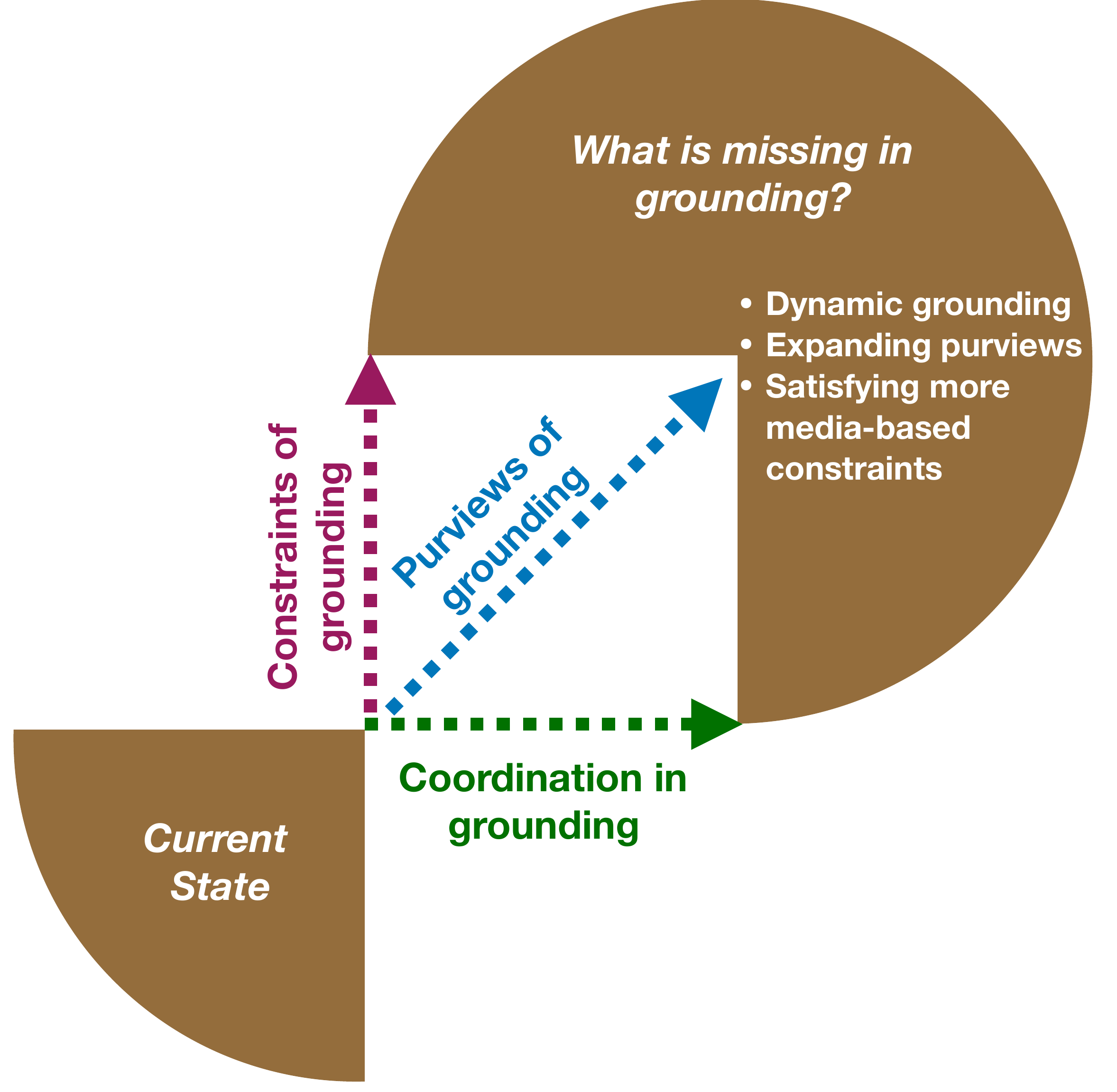}
\caption{Dimensions of grounding -- required to bridge the gap between current state of research and what is missing from a more complete sense of grounding. }
\label{fig:intro}
\end{figure}

\noindent In contrast, research in cognitive science defines grounding as the process of building a common ground based on shared mutual information in order to successfully communicate \cite{clark1982hearers, krauss1990mutual, DBLP:books/others/91/ClarkB91, lewis2008convention}. We argue that this definition subsumes NLP's current working definition and provides concrete guidance on which phenomena are missing to ensure the naturalness and long term utility of our technologies.

In Section \ref{sec:dimensions}, we formalize 3 dimensions key to grounding: Coordination, Purviews and Constraints, to systematize our analysis of limitations in current work. 
Section \ref{sec:grounding} presents a comprehensive review of the current progress in the field including the interplay of different domains, modalities, and techniques.
This analysis includes understanding when techniques have been specifically designed for a single modality, task, or form of grounding.
Finally, Section \ref{sec:repurpose} outlines strategies to repurpose existing datasets and tasks to align with the new richer definition from cognitive science literature. 
These introspections, re-formulations, and concrete steps situate NLP `grounding' in larger scientific discourse, to increase its relevance and promise.

\section{Dimensions of grounding}
\label{sec:dimensions}

Defining grounding loosely as \textit{linking} or tethering concepts is insufficient to achieve a more realistic sense of  grounding. Figure \ref{fig:intro} presents the research dimensions missing from most current work.

\subsection{Dimension 1: Coordination in grounding}
\label{sec:coordination}

\begin{figure}[t!]
\centering
\includegraphics[trim=3cm 6.5cm 1.8cm 4.8cm,clip,width=0.98\linewidth]{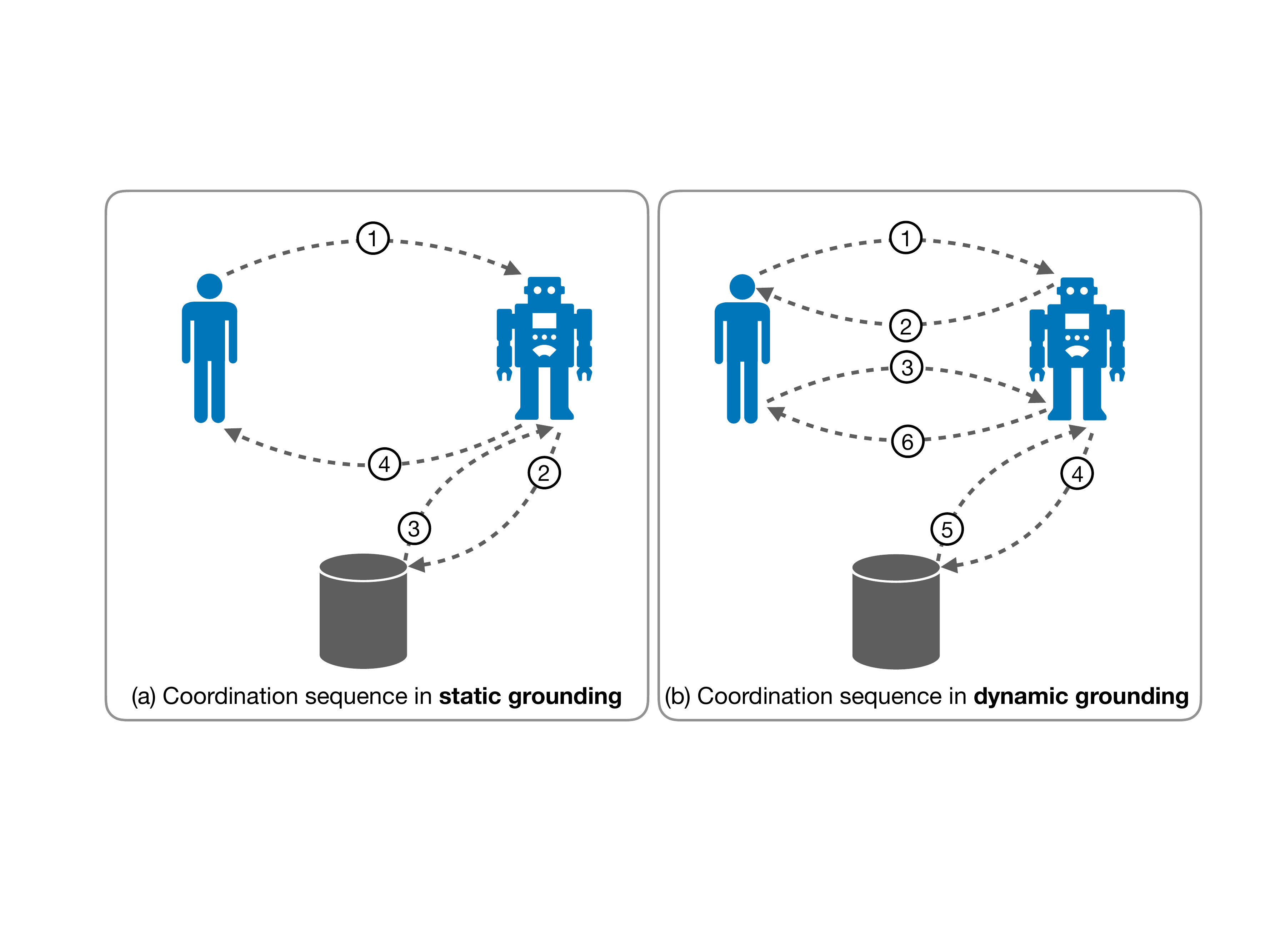}
\caption{Coordination sequence in grounding}
\label{fig:coordination}
\end{figure}

The first and the most important dimension that bridges the gap between the two definitions of grounding is the aspect of coordination -- alternatively viewed as the difference between \textit{static} and \textit{dynamic} grounding (Fig \ref{fig:coordination}). 

\paragraph{}{Static grounding} is the most common type and assumes that the evidence for common ground or the gold truth for grounding is given or attained pseudo-automatically. This is demonstrated in Figure \ref{fig:coordination} (a). The sequence for this form of interaction includes: (1) human querying the agent,  (2) agent querying the data or the knowledge it acquired, (3) agent retrieving and framing a response and (4) agent delivering it to the human. In this setting the common ground is the ground truth KB/data. The human and the agent have common ground by assuming its universality (i.e. no external references). Therefore, successfully grounding the query in this case relies solely on the agent being able to link the query to the data. For instance, in a scenario where a human wants to know the weather report, the accuracy of the database itself is axiomatic and we build a model for the agent to accurately retrieve the queried information in natural language. 
 
Most current research assumes static grounding so progress is measured by the ability of the agent to \textit{link} more concepts to more data. However, the axiomatic common ground often does not exist and needs to be established in real world scenarios.

\paragraph{Dynamic grounding } posits that common ground is built via interactions and clarifications. 
The mutual information needed to communicate successfully is built via 
interactions including: Requesting and providing clarifications, Acknowledging or confirming the clarifications, Enacting or demonstrating to receive confirmations, and so forth. 
This dynamically-established-grounding 
guides the rest of the interaction by course-correcting any misunderstandings. 
The sequence of actions in dynamic grounding is demonstrated in Figure \ref{fig:coordination} (b). The steps for establishing grounding is a part of the interaction that includes: (1) The human querying the agent, (2) The agent requesting clarification or acknowledging, (3) The human clarifying or confirming. These three steps loop until a common ground is established. The remaining steps of (4) querying the data, (5) retrieving or framing a response, and (6) delivering the response, are same as that of static grounding. 
The agent and the human may not be on the same common ground but steps 2 and 3 loop as the conversation progresses to build this common ground. 
The process of successfully grounding the query not only relies on the ability of the agent to \textit{link} the query but also 
to \textit{construct the common ground from the mutually shared information} with respect to the human. Although there are efforts about clarification questioning \cite{}, the coverage of phenomena are still far from comprehensive \cite{DBLP:journals/corr/abs-2104-08964}.

Cognitive sciences in the perspective of language acquisition \cite{carpenter1998social} present two ways of dynamic grounding via joint attention \cite{DBLP:conf/acl/KolevaVSK15, DBLP:conf/hri/TanABH20}: Dyadic joint attention and Triadic joint attention. In our case, dyadic attention describes the interaction between the human and the agent and any clarification or confirmation is done strictly between the both of them. Triadic attention also includes a tangible entity along with the human and the agent. The human can provide clarifications by gazing or pointing to this additional piece in the triad.

\begin{quote}
\begin{small}
\textit{Summary: 
The community should prioritize 
dynamic grounding 
as it is more general and more accurately matches real experiences.
}
\end{small}
\end{quote}





\subsection{Dimension 2: Purviews of grounding}
\label{sec:purviews}

Next, we present the different stages behind reaching a common ground, known as purviews. Most of the current approaches and tasks address these purviews individually and independently, while they are often co-dependent in real world scenarios.


\noindent \textbf{Stage 1: Localization: } 
The first stage is the localization of the concept either in the physical or mental contexts. This step is idiosyncratic and relates to the ability of the agent alone to localize the concept. These concepts often are also linked in a \textit{compositional} form. For instance, consider a scenario in which the agent is to locate a `blue sweater'. The agent needs to understand each of the concepts of `blue' and `sweater' individually and then locate the composition of the whole unit. 
\citet{clark2004speaking} from cognitive sciences demonstrate how \textit{incremental} grounding \cite{DBLP:conf/eacl/SchlangenS09, DBLP:conf/naacl/DeVaultT13, DBLP:conf/iwcs/EshghiHGHP15} is performed with these compositions and show how recognition and interpretation of fragments help in this by breaking down instructions into simpler ones. 
This localization occurs at word, phrase and even sentence level in the language modality and pixel, object and scene level in the visual modality.

\noindent \textbf{Stage 2: External Knowledge: } After localizing the concept, the next step is to ensure consistency of the current context of the concept with existing knowledge. 
Often times, the references of grounding either match or contradict the references from our prior knowledge and external knowledge. This might lead to misunderstandings in the consequent rounds of communication. Hence, in addition to localizing the concept, it is also essential to make the concept and its attributes consistent with the available knowledge sources. Most of the current research is focused on localizing with few efforts towards extending it to maintain a consistency of the grounded concept with other knowledge sources.



\noindent \textbf{Stage 3: Common sense: } After establishing consistency of the concept, a human-like interaction additionally calls for grounding the common sense associated with the concept in that scenario. In addition to the basic level of practical knowledge that concerns with day to day scenarios \citet{DBLP:conf/acl/SapSBCR20}, the concept should also be reasoned based on that particular context.
This contextual common sense moves the idiosyncratic sense towards a sense of collective understanding. For instance, if the human feels cold and asks the agent to get a blue coat, the agent needs to understand that the coat in this instance is a sweater coat and not a formal coat. This implicit common sense minimizes the effort in building a common ground reducing articulation of meticulous details. Therefore it is essential to incorporate this explicitly in our modeling as well.

\noindent \textbf{Stage 4: Personalized consensus: } As a part of the evolving conversations, the references in the language evolve as well. The grounded term might have different meanings for the agent in the context with access to the history as opposed to a fresh agent without access to the history. This multi-instance multi-turn process to achieve consensus makes this collective or a shared stage continually adapting to personalization leading to better engagement \cite{DBLP:conf/icmi/BohusH14}. In such settings, it is sufficient that the human and the agent are in consensus with the truth value of the grounded term, which need not be the same as the ground truth. This shift in the truth value of the meanings of the grounded terms often arise due to developing short-cuts for ease of communication and personalization, which is an acceptable shift as long as the communication is successful.

\begin{quote}
\begin{small}
\textit{Summary: Common ground requires expanding to verticals of local, general, common-sense and personalized contextual knowledge.
}
\end{small}
\end{quote}

\subsection{Dimension 3: Constraints of grounding}
\label{sec:constraints}

The medium and mode of communication constrain communicative goals in practical scenarios.
The number and availability of such media have increased and facilitated ubiquitous communication around the world, presenting a diversity in the mode of interaction. Motivated by this, we resurface and adapt the constraints of grounding with respect to media of interaction as defined by \citet{DBLP:books/others/91/ClarkB91}. Here are the definitions of these constraints in the context of grounded language processing and the corresponding categorization of the majority of the representative domains in grounding satisfying different constraints.

\noindent $\bigcdot$ \textit{Copresence: } Agent and human share the same physical environment of the data. Most of the current research in the category of embodied agents satisfy this constraint.

\noindent $\bigcdot$ \textit{Visibility: } The data is visible to the agent and/or human. The domains of images, images \& speech,	videos,	embodied agents satisfy this constraint.

\noindent $\bigcdot$ \textit{Audibility: } Agent and human communicate by speaking about the data. Domains like speech, spoken image captions and videos satisfy this.

\noindent $\bigcdot$ \textit{Cotemporality: } The agent/human receives at roughly the same time as the human/agent produces. The lag in the domains like conversations or interactive embodied agents is considered negligible and satisfy this constraint.

\noindent $\bigcdot$ \textit{Simultaneity: } The agent and the human can send and receive at once simultaneously. Most media are cotemporal but do not engage in simultaneous interaction. This often disrupts the understanding of the current utterance and the participant may have to repeat it to avoid misunderstandings, which is commonly observed in real world scenarios.

\noindent $\bigcdot$ \textit{Sequentiality: } The turn order of the agent and the human cannot get out of sequence. Face-to-face conversations usually follow this constraint but an email thread with active participants and the comments sections in online portals (such as Youtube, Twitch etc.,) do not necessarily follow a sequence. In such cases a reply to the message may be separated by arbitrary number of irrelevant messages. These categories are usually understudied but are commonly observed online.

\noindent $\bigcdot$ \textit{Reviewability: } The agent reviews the common ground to the human to adapt to imperfect human memories. For instance, we reiterate full references instead of adapting to short cut references when the conversation resurfaces after a while. This is to develop a personalized adaptation between the interlocutors based on the media to enable ease of communication.

\noindent $\bigcdot$ \textit{Revisability: } The interaction between the agent and the human indexes to a specific utterance in the conversation sequence and revise it, therefore changing the course of the interaction henceforth. Human errors are only natural in a conversation and the agent needs to be ready to rectify the previously grounded understanding.

There has been a good and continual effort in formulating tasks and datasets that satisfy the constraints of visibility, audibility and cotemporality. Contemporary efforts also see an increased interest in addressing copresence in grounded contexts. 
Very recently, \cite{DBLP:conf/eacl/BenottiB21} highlights the importance of recovering from mistakes while establishing the collabrative nature of grounding, contributing to the ability of revisability. 

\begin{quote}
\begin{small}
\textit{Summary: 
Key to progress is to focus on largely a blind spot in grounding:  simultaneity, sequentiality \& revisability to revive from mistakes.}
\end{small}
\end{quote}




\section{Grounding `\textit{Grounding}'}
\label{sec:grounding}

Having covered a more formal definition of grounding adapted to NLP, we turn our attention to cataloging the precise usage of `grounding' in our research community.
We present an analysis on the various domains and techniques NLP has explored. 

\subsection{Data and Annotations}


To this end, since our aim is to investigate how the community understands the loosely defined term `grounding', we subselected all the papers that mention terms for `grounding' in the title or abstract 
from the S2ORC data \cite{lo-wang-2020-s2orc} between the years 1980-2020. In this way, we grounded the term \textit{`grounding'} in literature \footnote{Please note that this is not an exhaustive list of papers working on grounding as there are several others that do mention this term and still work on some form of grounding.} to collect the relevant papers. 
We acknowledge that the papers analyzed here are not exhaustive with respect to concept of `grounding'.

Each of the paper is annotated with answers to the following questions: (i) is it introducing a new task? (ii) is it introducing a new dataset? (iii) what is the world scope (iv) is it working on multiple languages? (v) what are the grounding domains? (vi) what is the grounding task? (vii) what is the grounding technique?

\subsection{Domains of grounding}

Real world contexts we interact with are diverse and can be derived from different modalities such as textual or non-textual, each of which comprises of domains.
Our categorization of these is inspired from the constraints of grounding as described in \S \ref{sec:constraints}.
Based on this, the modality based categorization include the following domains:

\noindent $\bigcdot$ \textit{Textual modality  comprising plain text, entities \& events, knowledge bases and knowledge graphs.}

\noindent $\bigcdot$ \textit{Non-textual modality comprising images, speech, images \& speech and videos.}

Numerous other domains including \textit{numbers and equations, colors, programs, tables, brain activity signals} etc., are studied in the context of grounding at relatively lower scale in comparison to the aforementioned ones.
Each of these can further be interacted with along the variation in the coordination dimension of grounding from \S \ref{sec:coordination}, that give rise to the following settings including \textit{conversations, embodied agents and face-to-face interactions}.

\subsection{Approaches to grounding}
\label{sec:approaches}

\begin{figure}[t!]
\centering
\includegraphics[trim=4cm 8cm 3cm 3cm,clip,width=0.99\linewidth]{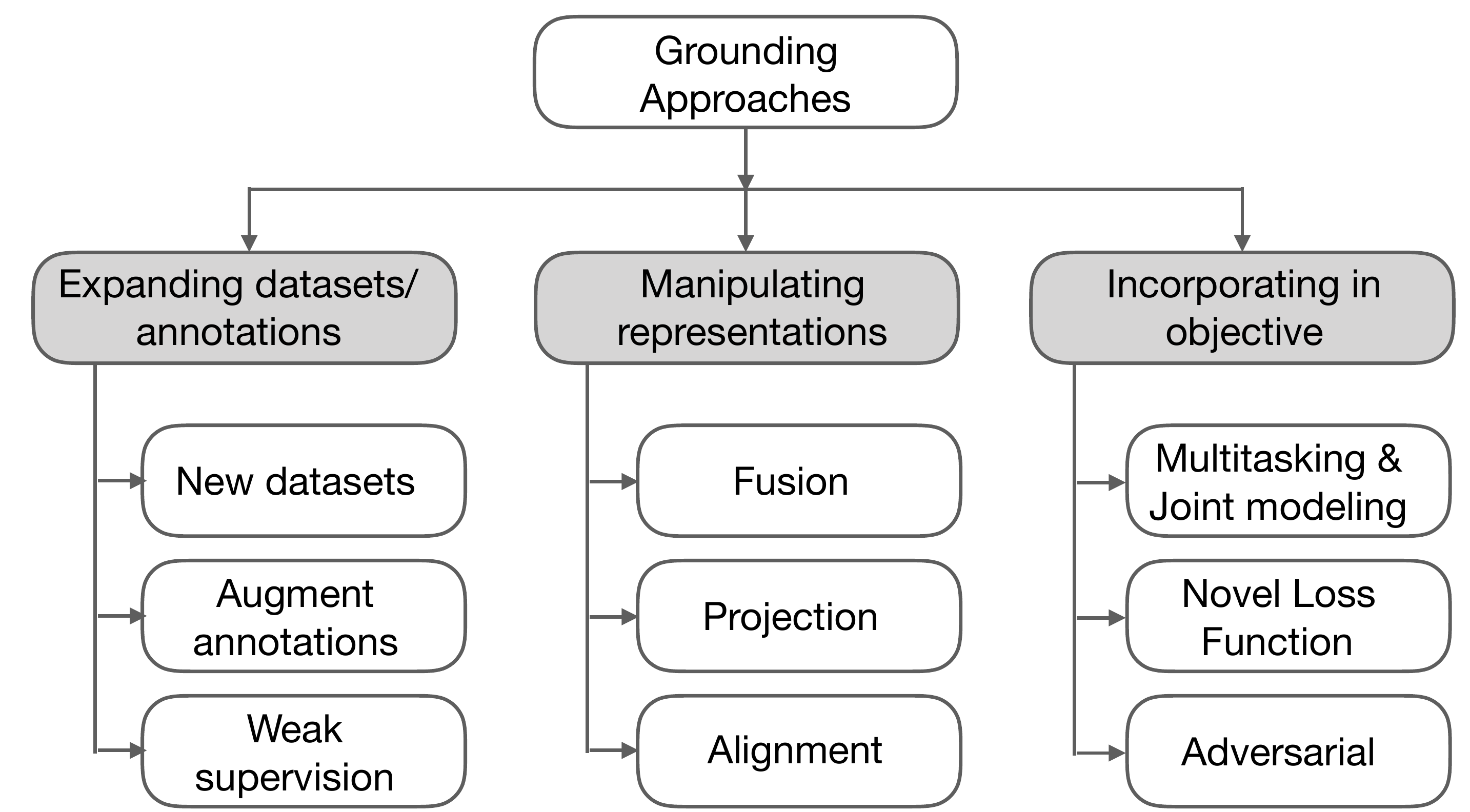}
\caption{Categorical approaches to grounding }
\label{fig:techniques}
\end{figure}

This section presents a list of approaches tailored to grounding. The obvious solution is to expand the datasets to promote a research platform. The second is to manipulate different representations to \textit{link} and bring them together. Finally the learning objective can leverage grounding. The sub-categories within each are presented in Figure \ref{fig:techniques}.


\paragraph{1. Expanding datasets / annotations: } The first step towards building an ecosystem for research in grounding is to  curate the necessary datasets which is accomplished with expensive human efforts, augmenting existing annotations and automatically deriving annotations with weak supervision.

\noindent \textbf{1a) New datasets:} There has been an increase in efforts for curating new datasets with task specific annotations. These are briefly overlaid in Table \ref{tab:new-datasets} along with their modalities, domains and tasks.

\noindent \textbf{1b) Augment annotations:} These curated datasets can also be used subsequently to augment with task specific annotations instead of collecting the data from scratch, which might be more expensive.

\noindent \textit{ $\bigcdot$ Non-textual Modality:}
Static grounding here includes using adversarial references to ground visual referring expressions \cite{DBLP:conf/acl/AkulaGAZR20}, narration \cite{DBLP:conf/acl/ChanduNB19, DBLP:conf/emnlp/ChanduDB20},  
language learning \cite{DBLP:conf/acl/SugliaKVBEFL20, DBLP:conf/emnlp/JinDSNR20} etc., 

\noindent \textit{ $\bigcdot$ Textual Modality: }
Static grounding includes entity slot filling  \cite{DBLP:conf/emnlp/BiskRBHS16}.

\noindent \textit{ $\bigcdot$ Interactive: }
Though not fully dynamic grounding, some efforts here are amongst tasks like understanding spatial expressions \cite{DBLP:conf/emnlp/UdagawaYA20}, 
collaborative drawing \cite{DBLP:conf/acl/KimKCRZTBP19} etc.,

\noindent \textbf{1c) Weak supervision: }
While the above two are based on human efforts, we can also perform weak supervision to use a model trained to derive automatic soft annotations required for the task.


\begin{table}[t]
\centering
\resizebox{0.48\textwidth}{!}{%
\begin{tabular}{@{}c@{\hspace{5pt}}l@{\hspace{5pt}}l@{\hspace{5pt}}l@{}}
\toprule
\textbf{Modality} & \textbf{Domain} & \textbf{Task} & \textbf{Work} \\
\midrule
\multirow{16}{*}{\rotatebox{90}{Non-textual}} & \multirow{12}{*}{Images} & caption relevance & \cite{DBLP:conf/acl/SuhrZZZBA19} \\ 
 &  & multimodal MT & \cite{DBLP:conf/emnlp/ZhouCLY18} \\ 
 &  & sports commentaries & \cite{DBLP:conf/emnlp/Koncel-KedziorskiHF14} \\ 
 &  & semantic role labeling & \cite{DBLP:conf/emnlp/SilbererP18} \\ 
 &  & instruction following & \cite{DBLP:conf/eacl/SchlangenH17a} \\ 
 &  & navigation & \cite{DBLP:conf/conll/AndreasK14} \\ 
 &  & causality & \cite{DBLP:conf/acl/GaoDYC16} \\ 
 &  & spatial expressions & \cite{DBLP:conf/acl/KelleherKC06} \\ 
 &  & spoken image captioning & \cite{DBLP:conf/conll/AlishahiBC17} \\ 
 &  & entailment & \cite{DBLP:conf/coling/Vu0EJLTTBG18} \\ 
 &  & image search & \cite{DBLP:conf/acl/KirosCH18} \\ 
 &  & scene generation & \cite{DBLP:conf/acl/ChangMSPM15} \\ \cline{2-4} 
 & \multirow{4}{*}{Videos} & action segmentation & \cite{DBLP:journals/tacl/RegneriRWTSP13} \\ 
 &  & semantic parsing & \cite{DBLP:conf/emnlp/RossBBMK18} \\ 
 &  & instruction following & \cite{DBLP:conf/emnlp/LiuYSSHZC16} \\ 
 &  & question answering & \cite{DBLP:conf/acl/LeiYBB20} \\ \midrule 
\multirow{6}{*}{\rotatebox{90}{Textual}} & \multirow{6}{*}{Text} & content transfer & \cite{DBLP:conf/naacl/PrabhumoyeQG19} \\ 
 &  & commonsense inference & \cite{DBLP:conf/emnlp/ZellersBSC18} \\ 
 &  & reference resolution & \cite{DBLP:conf/acl/KenningtonS15} \\ 
 &  & symbol grounding & \cite{DBLP:conf/emnlp/KamekoMT15} \\ 
 &  & bilingual lexicon extraction & \cite{DBLP:conf/coling/LawsMDSHS10} \\ 
 &  & POS tagging & \cite{DBLP:conf/naacl/CardenasLJM19} \\ 
 \midrule 
\multirow{10}{*}{\rotatebox{90}{Interactive}} & \multirow{3}{*}{Text} & negotiations & \cite{DBLP:conf/emnlp/CadilhacABL13} \\ 
 &  & documents & \cite{DBLP:conf/emnlp/ZhouPB18} \\ 
 &  & improvisation & \cite{DBLP:conf/acl/ChoM20} \\ \cline{2-4} 
 & \multirow{6}{*}{Visual} & \multirow{2}{*}{referring expressions} & \cite{DBLP:conf/acl/HaberBTGBF19} \\ 
 &  &  & \cite{takmaz2020refer} \\ 
 &  & emotions and styles & \cite{DBLP:conf/acl/ShusterHBW20} \\ 
 &  & media interviews & \cite{DBLP:conf/emnlp/MajumderLNM20} \\ 
 &  & spatial reasoning & \cite{DBLP:journals/tacl/JannerNB18} \\ 
 &  & navigation & \cite{DBLP:conf/emnlp/KuAPIB20} \\ \cline{2-4} 
 & Other & problem solving & \cite{DBLP:conf/naacl/LiB15} \\ 
 \bottomrule
\end{tabular}%
}
\caption{Example datasets introduced for \textit{grounding}.}
\label{tab:new-datasets}
\end{table}

\noindent \textit{ $\bigcdot$ Non-Textual Modality: }
In the visual modality, weak supervision is used in the contexts of automatic object proposals for different tasks like spoken image captioning \cite{DBLP:conf/emnlp/SrinivasanSME20}, visual semantic role labeling \cite{DBLP:conf/emnlp/SilbererP18}, phrase grounding \cite{DBLP:conf/acl/ChenMLW19}, 
loose temporal alignments between utterances and a set of events \cite{DBLP:conf/emnlp/Koncel-KedziorskiHF14} etc.,

\noindent \textit{ $\bigcdot$ Textual Modality: }
In the contexts of text, \citet{DBLP:journals/tacl/TsaiR16} work towards disambiguating concept mentions appearing in documents and grounding them in multiple KBs which is a step towards Stage 3 in \S \ref{sec:purviews}. 
\citet{DBLP:conf/acl/Poon13} perform question answering with a single database and \cite{DBLP:conf/naacl/ParikhPT15} with symbols.

\begin{quote}
\begin{small}
\textit{Summary: While augmentation and weak supervision can be leveraged for dimensions of coordination and purviews, curating new datasets is the need of the hour to explore various constraints.}
\end{small}
\end{quote}


\paragraph{2. Manipulating representations:} Grounding concepts often involves multiple modalities or representations that are \textit{linked}. Three major methods to approach this are detailed here.

\noindent \textbf{2a) Fusion and concatenation:} Fusion is a very common technique in scenarios involving multiple modalities. In scenarios with a single modality, representations are often concatenated.

\noindent \textit{ $\bigcdot$ Non-textual modality: }
Fusion is applied with images for tasks like referring expressions \cite{DBLP:conf/conll/RoyNPPR19}, SRL  \cite{DBLP:conf/naacl/YangGLXZC16} etc.,
For videos, some tasks are grounding action descriptions \cite{DBLP:journals/tacl/RegneriRWTSP13}, 
spatio-temporal QA \cite{DBLP:conf/acl/LeiYBB20}, concept similarity \cite{DBLP:conf/emnlp/KielaC15}, mapping events 
\cite{DBLP:conf/acl/FleischmanR08} etc.,

\noindent \textit{ $\bigcdot$ Textual Modality: }
With text, this is similar to concatenating context (\citet{DBLP:conf/naacl/PrabhumoyeQG19} perform content transfer by augmenting context).

\noindent \textit{ $\bigcdot$ Interactive: }
In a conversational setting, work is explored in reference  resolution \cite{takmaz2020refer, DBLP:conf/acl/HaberBTGBF19}, 
generating engaging response \cite{DBLP:conf/acl/ShusterHBW20},  
document grounded response generation \citet{DBLP:conf/emnlp/ZhouPB18}, etc.,

\noindent \textit{ $\bigcdot$ Others:}
\citet{DBLP:conf/acl/NakanoRSC03} study face-to-face grounding in instruction giving for agents.

\noindent \textbf{2b) Alignment:} An alternative to combining representations is aligning them with one another.

\noindent \textit{ $\bigcdot$ Non-textual modality: }
\citet{DBLP:conf/emnlp/WangTSMY20} perform phrase localization in images and \citet{DBLP:conf/emnlp/HesselZPS20} study temporal alignment in videos.

\noindent \textit{ $\bigcdot$ Interactive: }
\citet{DBLP:conf/eacl/SchlangenH17a} align GUI actions to sub-utterances in conversations and \citet{DBLP:journals/tacl/JannerNB18} align local neighborhoods to the corresponding verbalizations. 

\noindent \textbf{2c) Projecting into a common space:} A widely used approach is to also bring the different representations on to a joint common space.

\noindent \textit{ $\bigcdot$ Non-textual modality: }
Projection to a joint semantic space is used in 
spoken image captioning \cite{DBLP:conf/acl/ChrupalaGA17, DBLP:conf/conll/AlishahiBC17,  DBLP:conf/conll/HavardCB19}, bicoding for learning image attributes \cite{DBLP:conf/acl/SilbererL14}, representation learning of images \cite{DBLP:conf/emnlp/ZarriessS17} and speech \cite{DBLP:conf/emnlp/VijayakumarVP17}.

\noindent \textit{ $\bigcdot$ Textual modality: }
\citet{DBLP:conf/coling/TsaiR16} demonstrate cross-lingual NER and mention grounding model by activating corresponding language features.
\citet{DBLP:conf/acl/YangZSD19} perform imputation of embeddings for rare and unseen words by projecting a graph to the pre-trained embeddings space.

\begin{quote}
\begin{small}
\textit{Summary: Modeling different representations effectively aid in improving both consistency across purviews and media based constraints.}
\end{small}
\end{quote}


\paragraph{3. Learning Objective:} Grounding is often performed to support a more defined end purpose task. 
We identified 3 ways that are broadly adopted to incorporate grounding in objective functions. 


\noindent \textbf{3a) Multitasking and Joint Modeling: }The \textit{linking formulation} of grounding is often used as an auxiliary or dependent to model another task.

\noindent \textit{ $\bigcdot$ Non-textual Modality: }
Multitasking with images is used to perform spoken image captioning \cite{DBLP:conf/acl/Chrupala19} and grammar induction \cite{DBLP:conf/emnlp/ZhaoT20}. Joint modeling was used in multi-resolution language grounding \citet{DBLP:conf/emnlp/Koncel-KedziorskiHF14}, identifying referring expressions \citet{DBLP:conf/conll/RoyNPPR19}, multimodal MT \cite{DBLP:conf/emnlp/ZhouCLY18}, video parsing  \citet{DBLP:conf/emnlp/RossBBMK18}, learning latent semantic annotations \cite{DBLP:conf/emnlp/QinYWWL18} etc.,


\begin{figure*}[t!]
\centering
\includegraphics[trim=0.5cm 1.5cm 0cm 2.0cm,clip,width=0.75\linewidth]{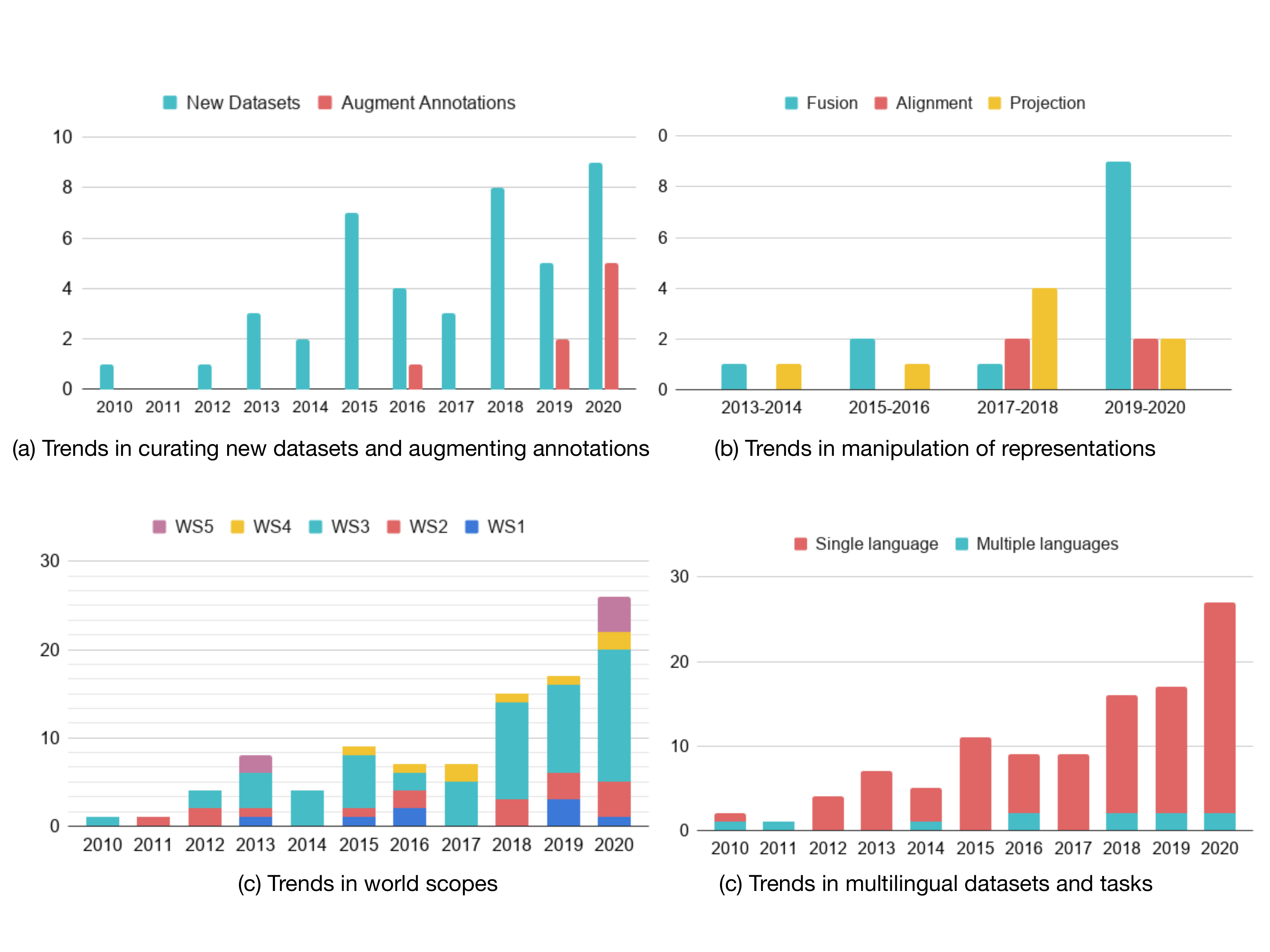}
\caption{Analysis on the trends in grounding}
\label{fig:trends}
\end{figure*}

\noindent \textit{ $\bigcdot$ Interactive: }
In a conversational setting, multitasking is used to compute concept similarity judgements \cite{DBLP:conf/acl/SilbererL14}, knowledge grounded response generation \cite{DBLP:conf/emnlp/MajumderLNM20}, grounding language instructions \citet{DBLP:conf/acl/HuFRKDS19}. Joint modeling is used by \citet{DBLP:conf/naacl/LiB15} to address dialog for complex problem solving in computer programs.

\noindent \textbf{3b) Loss Function: } It is crucial to utilize appropriate loss designed for the specific grounding task. The main difference between multitasking and a loss function adaptation is that while multitasking reweights combinations of existing loss functions, novel loss functions are informed by the data/task at hand, adapting to a novel use case.

\noindent \textit{ $\bigcdot$ Non-textual Modality: }
\citet{DBLP:conf/conll/GrujicicRTB20} design soft organ distance loss to model 
inter and intra organ interactions for relative grounding. \citet{DBLP:conf/conll/IlharcoZB19} improve diversity in spoken captions with a masked margin softmax loss.


\noindent \textbf{3c) Adversarial:} Leveraging deceptive grounded inputs in an attempt to fool the model is capable of making it robust to certain errors.

\noindent \textit{ $\bigcdot$ Non-textual Modality: }
\citet{DBLP:conf/acl/HsiehYCZC18, DBLP:conf/acl/AkulaGAZR20} present an algorithm to craft visually-similar adversarial examples.

\noindent \textit{ $\bigcdot$ Textual Modality: }
\citet{DBLP:conf/emnlp/ZellersBSC18} perform adversarial filtering and constructs a de-biased dataset by iteratively training 
stylistic classifiers.


\begin{quote}
\begin{small}
\textit{Summary: Manipulating learning objective is a modeling capability aiding as an additional component in bringing grounding adjunct to several other end tasks across all the dimensions.}
\end{small}
\end{quote}

\subsection{Analysis of trends}
\label{sec:analysis}

Based on the categories of approaches and different datasets from \S \ref{sec:approaches}, 
we presented a representative set of analyses that highlight the major avenues that addressing the key missing pieces of work on grounding to advance future research.


Figure \ref{fig:trends} presents the trends in the development of grounding over the past decade including: specific approaches (a,b) that presents new tasks/challenges; world scopes \cite{DBLP:conf/emnlp/BiskHTABCLLMNPT20} (c) contributing to grounding language in different data types; and multilinguality (d) contributing to a part of linguistic diversity.
We also present hierarchical pie charts in Figure \ref{fig:analysis_graphs} and in Appendix to analyze the compositions of modalities and domains for these approaches.
While we believe our analysis targets several of the most critical dimensions paving way for future research directions, it is not exhaustive and welcome suggestions from the community for additional analysis. For example, it is also interesting to study domain diversity, task formulation/usefulness, etc., in future. 

\begin{figure*}[t!]
\centering
\includegraphics[trim=0.8cm 5cm 0.7cm 4cm,clip,width=0.72\linewidth]{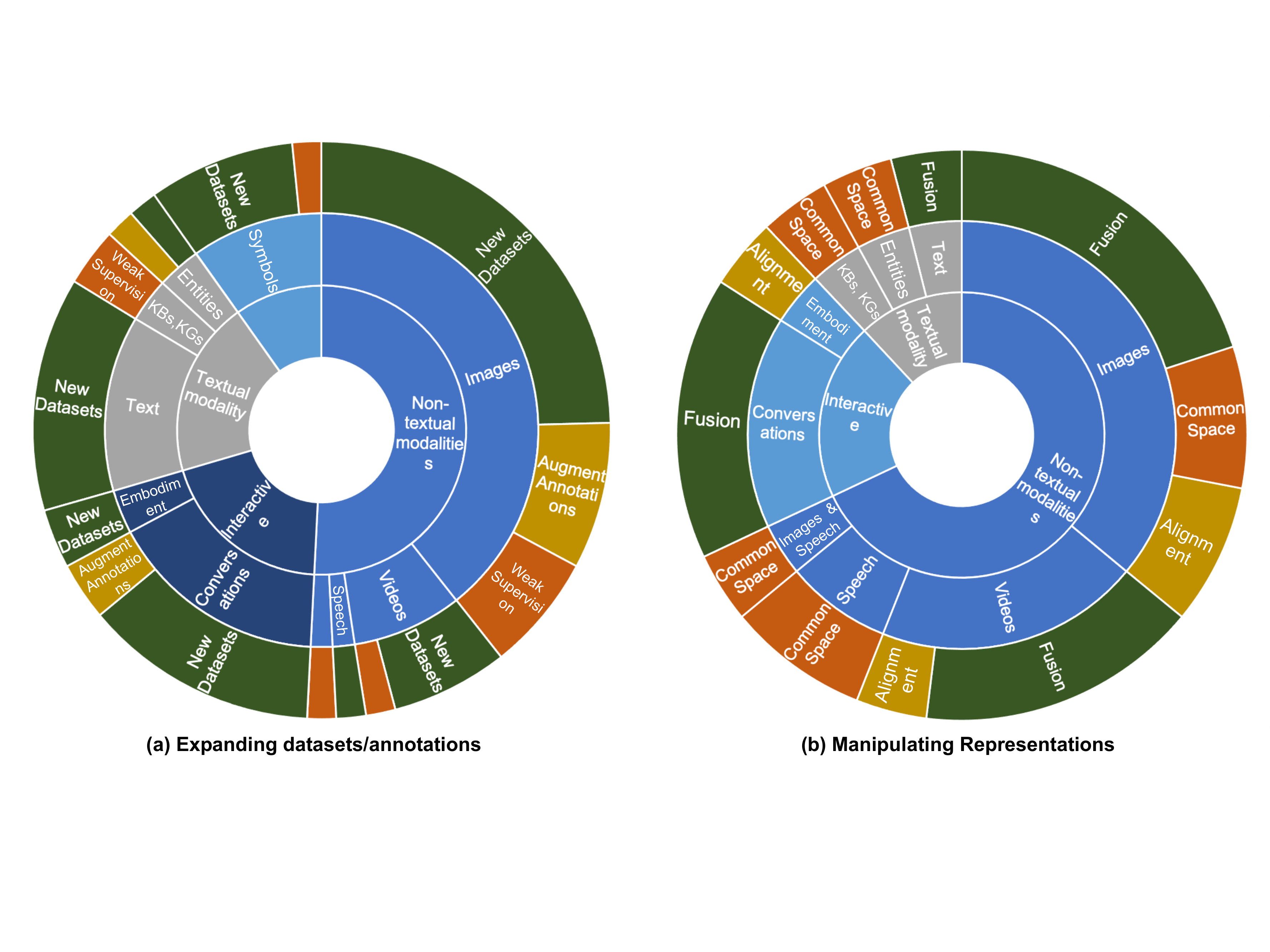}
\caption{Analysis of Domains and Techniques}
\label{fig:analysis_graphs}
\end{figure*}

\noindent \textbf{Trends in datasets expansion: }The introduction of new datasets has seen a rapid increase over the years, while there is also a subtle increasing trend in augmenting annotations to the existing datasets, as observed in Figure \ref{fig:trends} (a). As we can see from Figure \ref{fig:analysis_graphs} (a), across all the domains, gathering new datasets seem to be prominent than augmenting them with additional annotations to repurpose the data for a new task. There seems to be a higher emphasis of expansion of datasets in the non-textual modalities, particularly in the domain of images. A similar rise is not observed in interactive settings including conversational data and interaction with embodied agents; which is the propitious way to bridge the gap towards real sense of grounding. It is indeed encouraging to see an increasing trend in the efforts for expanding datasets but \textit{the need of the hour is to redirect some of these resources to address dynamic grounding in the coordination dimension which is scarcely studied in existing datatsets}.

\noindent \textbf{Trends in manipulating representations: } From Figure \ref{fig:trends} (b), we note that the fusion technique has and is increasingly becoming popular in grounding through manipulating representations in comparison to alignment and projection. This is also observed in Figure \ref{fig:analysis_graphs} (b) with the dominance of non-textual modality. In the context of textual modality, this technique is equivalent to concatenation of the context or history in a conversation. Projecting onto a common space is the next popular technique in comparison to alignment. Similarly, we observe that the non-textual modality overwhelmingly occupies the space of manipulating representations with exceeding prominence of fusion. \textit{Fusion and projecting onto common space currently are exceedingly used methodologies to ground within a single purview. They demonstrate a promising direction to manipulate representations across different stages to maintain consistency along the purviews.}

\noindent \textbf{Trends in World Scopes: }We also study the development of the field based on the definitions of the world scopes presented by \citet{DBLP:conf/emnlp/BiskHTABCLLMNPT20}. Based on this, last decade has seen an increasing dominance in research on world scope 3 (world of sights and sounds). However, this is limited to this scope and the same trend is not clear in world scope 4 (world of embodiment and action). An encouraging observation is the focus of the field in world scope 5 (social world) which is closer to real interactions in the last year. \textit{We need to accelerate development of datasets and tasks in world scopes 4 and 5. It is highly recommended to take dynamic grounding scenario into account in the efforts for curating  datasets in these scopes.}

\noindent \textbf{Inclusivity of multiple languages: } Figure \ref{fig:trends} (c) shows that research into grounding in multiple languages is still incredibly rare. As noted by \citet{bender2011achieving}, improvements in one language do not necessarily mandate comparable performances in other languages. 
\textit{ The norm for benchmarking large scale tasks still remains anglo-centric and we need serious efforts to drift this trend to identify challenges in grounding across languages. As a first step, a relatively less expensive way to navigate this dearth is to augment the annotations of existing datasets with other languages.}

\section{Path Ahead: Towards New Tasks and Repurposing Existing Datasets}
\label{sec:repurpose}

We presented the dimensions of grounding that require serious attention to bridge the gap between the definitions in cognitive sciences and language processing communities in \S \ref{sec:dimensions}. Based on this, we analyzed the language processing research to understand where we stand and where we fall short with the ongoing efforts in trends in grounding in \S \ref{sec:grounding}. While we strongly advocate for efforts in building new datasets and tasks considering progress along these dimensions, we believe in a smoother transition towards this goal. Hence we present strategies to repurpose existing resources to maximum utility as we stride towards achieving grounding in real sense. In this section, we focus on concrete suggestions to improve along each of the dimensions.

\noindent \textbf{Coordination: } This is based on simulating interaction for dynamic grounding. As establishing a common ground is not integrated within  datasets, we propose an iterative paradigm to explicitly settle on a common ground based on our priors. 

The first family of methods to perform this is human-in-the-loop interactions. The traditional methods of data collection do not cater to human feedback or generation. Some recent approaches to incorporate human feedback are during data collection \cite{wallace2019trick}, training \cite{DBLP:journals/corr/abs-2009-01325}, inference \cite{DBLP:conf/acl/HancockBMW19}. While the feedback in a human in the loop setting can be via scores, we argue for natural language feedback \cite{wallace2019trick} loop, which resembles human-human grounding via communication.

The second family of methods are inspired from the theory of mind \cite{gopnik1992child} to iteratively or progressively ask and clarify to establish a common ground \cite{roman2020rmm}.
\citet{DBLP:conf/cvpr/VriesSCPLC17, DBLP:conf/acl/SugliaKVBEFL20} disambiguate or clarify the referenced object through a series of questions in a guessing game.
This iterative paradigm can be related to work by \citet{DBLP:conf/emnlp/ShwartzWBBC20} that generates clarification questions and answers to incorporate in the task of question answering. 
This loop of semi-automatic generation of clarifications establishes a common ground. 
This is also in spirit similar to generating an explanation or a hypothesis for question answering \cite{DBLP:journals/corr/abs-2004-05569}. The process of generating an acceptable explanation to human before acts as establishing a common ground.

We believe that datasets and tasks along the following three directions encourage dynamic grounding: (1) conversational language learning \cite{DBLP:conf/iclr/Chevalier-Boisvert19} or acquisition, and (2) clarification questioning and ambiguity resolution \cite{DBLP:conf/emnlp/ShwartzWBBC20} (3) mixed initiative for grounding in conversations \cite{DBLP:conf/sigdial/MorbiniFDSTR12}.

The need of the hour that can revolutionize this paradigm is the \textit{development of evaluation strategies to monitor evolution of the common ground.} This dynamic grounding data helps improve performance/robustness and encourages human's trust while using these interactive systems.

\noindent \textbf{Purviews: } This is based on establishing consistency across stages of grounding with an incremental paradigm. A simple solution is a modular approach where the purviews flow into the next stage after reasonably satisfying the previous stage. 
The current benchmarking approaches are mostly lateral i.e., our current strategies collate multiple datasets of a single task to benchmark. This approach implicitly establishes boundaries between the purviews. In contrast, \textit{we advocate for a longitudinal approach for benchmarking} i.e in addition to collating different datasets for a task, we also extend the purviews of the task such that the output from the previous purview flows into the next purview. An example of establishing a longitudinal benchmark for  visual dialog. The tasks flow from object detection (stage 1: localization) to knowledge graphs (stage 2: external knowledge) to common sense understanding (stage 3: common sense) to  empathetic dialogue (stage 4: personalization) for the \textit{same} dataset. This helps us dissect which aspect of grounding is the model good and bad at to understand the weak areas. 



\noindent \textbf{Constraints: } With media imposed constraints, there is a need for paradigm shift in the way these datasets are curated. The optimal way to navigate this problem is curating new datasets to \textit{specifically focus on the less studied  constraints of simultaneity, sequentiality and revisability.} At the heart of revisability in a collaborative dialog is clarification questioning and resolving ambiguities \cite{DBLP:conf/naacl/BoniM03, DBLP:conf/acl/DaumeR18, DBLP:conf/chiir/BraslavskiSAD17, DBLP:conf/acl/KumarB20, DBLP:journals/corr/abs-2009-11352, DBLP:journals/corr/abs-2104-08964}
However, they are rarely explored and are not systematically standardized across modalities. Transferring knowledge for shared constraints across tasks is a promising way to leverage the existing datasets.

\noindent \textbf{Augment with multilingual annotations: } Different languages also bring novel challenges to each of these issues (e.g. pronoun drop dialogue in Japanese, morphological alignments, etc). However, as observed in \S \ref{sec:analysis}, the increase in expanding datasets is not proportionally reflected to include multiple languages. We recommend a relatively less expensive process of translating the datasets for grounding into other languages to kick start this inclusion. The research community has already seen such efforts in image captioning with human annotated German captions in Multi30k \cite{DBLP:conf/acl/ElliottFSS16} extended from Flick30k \cite{DBLP:conf/iccv/PlummerWCCHL15} and  Japanese captions in STAIR  \cite{DBLP:conf/acl/YoshikawaST17} based on MS-COCO images \cite{DBLP:conf/eccv/LinMBHPRDZ14}. Instead of using human annotations, some efforts have also been made to use automatic translations such as the work by \citet{DBLP:conf/acl/ThapliyalS20} and denoising \cite{chandu2020weakly} extending from \cite{DBLP:conf/acl/SoricutDSG18}. Not just augmentation, but there are also ongoing efforts in gathering datasets in multiple languages \cite{DBLP:conf/emnlp/KuAPIB20} extending \cite{DBLP:conf/cvpr/AndersonWTB0S0G18}.


\section{Conclusions}
We discussed the missing pieces and dimensions that bridge the gap between the definitions of grounding in Cognitive Sciences and NLP communities. Thereby, we chart out executable actions in steering existing resources to bridge this gap along these dimensions to achieve a more realistic sense of grounding. Specifically: (1) Static grounding still remains the central tenet for existing tasks and datasets. However, 
dynamic grounding is key moving forward.
(2) Current benchmarking strategies evaluate model generalization. 
In tandem, we also need to steer towards longitudinal benchmarking to naturally proliferate across purviews of grounding that is closer to natural human interactions.
(3) Constraints imposed by the medium of communication present nuanced categories of communicative goals. While discerning learning from shared constraints, we also urge the community to invest resources on revisability as a way to recover from contextually mistaken groundings. 
While ruminating on the above phenomena, the challenge of expanding them to multiple languages and domains still persists.
We also recommend systematic evaluation of grounding along these dimensions in addition to the existing linking capabilities.

\section*{Ethical Considerations}
The analytical and ontological discussion here focuses exclusively on the question of \textit{grounding} and common ground and does not address the harmful biases inherent in these datasets. Further, the common ground for which we are advocating is culturally specific and future work that introduces tasks and data for these purposes must be explicit about who they serve (culturally and linguistically).

\bibliographystyle{acl_natbib}
\bibliography{anthology,acl2021}

\newpage
\pagebreak
\clearpage

\appendix

\section{Examples for dimensions of grounding}

\paragraph{Static Grounding: }
In static grounding, when you ask an agent \textit{``Can you place the dragon fruit on the rack''?}, the agent links the entities and places the dragon fruit on the rack. The challenge here is mainly the linking part which is crucial to ensure it accurately understood the instruction.

\paragraph{Dynamic Grounding: }
The same is not true for dynamic grounding. There are primarily 2 ways to materialize this. 
First, with respect to language learning:
What if the agent does not know dragon fruit?
The agent needs to first ask \textit{``What is a dragon fruit?''}, and the human provides an answer.
Lets say the human responded by describing the physical attributes such as \textit{reddish pink fruit} and/or a spatial reference by refering to it as \textit{the fruit on the bottom left}. The important aspect here is that the agent asks and learns what a dragon fruit is and use this knowledge later.

The second is ambiguity resolution.
Consider a scenario where there are multiple racks. It is very natural for a human to ask the agent which rack to resolve ambiguity.We expect the same from the agent to ask a clarifying question to resolve ambiguity and then place it on the second rack.

\paragraph{Purviews - Localization: }
Consider this example of a conversation between an agent and a human.
\vspace{1em}
\hrule
\vspace{0.5em}
\noindent \textit{\textbf{Human}: What is the name of the role Robert Downey Jr played in Avengers?}

\noindent \textit{\textbf{Agent}: He played the role of   Tony Stark, and sometimes is also referred to as 
Iron Man.}
\vspace{0.5em}
\hrule
\vspace{1em}
The agent begins by localizing and linking Robert Downey Jr to Tony Stark and Iron Man to provide the appropriate answer to the query. 

\paragraph{Purviews - External Knowledge: }
However, natural conversations also extend beyond the purview of localization to discuss a broadened scope involving external knowledge of the context including entities, actions etc., For example, consider this conversation which seems to be a natural continuation to the earlier one.
\vspace{1em}
\hrule
\vspace{0.5em}
\noindent \textit{\textbf{Human}: Is he the head of SHIELD?}

\noindent \textit{\textbf{Agent}: Tony Stark has never been the head of SHIELD in the movies but has been the acting head upon Maria Hill’s suggestion in the Comics.}
\vspace{0.5em}
\hrule
\vspace{1em}
Once we localized Tony Stark, asking additional information like whether he is the head of SHIELD is natural in conversations; However, access to required external knowledge is rarely present in the datasets as well as evaluated. Here, we need to refer to external sources spanning from movies to comics to conclude that he has been the acting head in the comics but was never in the movies.

\paragraph{Purviews - Common sense: }
One of the branches of natural progression to this context can extend to the following turns: 

\vspace{1em}
\hrule
\vspace{0.5em}
\noindent \textit{\textbf{Human}: How long was the contract between Tony Stark and Marvel?}

\noindent \textit{\textbf{Agent}: Tony Stark is the name of the character in Marvel. Would you like to know the 
contract length for Robert Downey Jr who played the role?}
\vspace{0.5em}
\hrule
\vspace{1em}

Here, the agent needs to understand that Tony Stark is not a real person, but is a character in Marvel. Hence, any contract is with the actor but not the character who played the role. The agent needs to have the common sense to understand this and clarify the question. 

\begin{table*}[tbh]
\centering
\resizebox{\textwidth}{!}{%
\begin{tabular}{l|ccc|ccccc}
\hline
 & \multicolumn{3}{c|}{\textbf{Modality}} & \multicolumn{5}{c}{\textbf{Cue}} \\
 & \multicolumn{1}{l}{\textbf{Copresence}} & \multicolumn{1}{l}{\textbf{Visibility}} & \multicolumn{1}{l|}{\textbf{Audibility}} & \multicolumn{1}{l}{\textbf{Cotemporality}} & \multicolumn{1}{l}{\textbf{Simultaneity}} & \multicolumn{1}{l}{\textbf{Sequentiality}} & \multicolumn{1}{l}{\textbf{Reviewability}} & \multicolumn{1}{l}{\textbf{Revisabiility}} \\ \hline
Face-to-face & \CheckmarkBold & \CheckmarkBold & \CheckmarkBold & \CheckmarkBold & \CheckmarkBold & \CheckmarkBold &  &  \\
Telephone &  &  & \CheckmarkBold & \CheckmarkBold & \CheckmarkBold & \CheckmarkBold &  &  \\
Video Teleconference &  & \CheckmarkBold & \CheckmarkBold & \CheckmarkBold & \CheckmarkBold & \CheckmarkBold &  &  \\
Terminal Teleconference &  &  &  & \CheckmarkBold &  & \CheckmarkBold & \CheckmarkBold &  \\
Answering Machines &  &  & \CheckmarkBold &  &  &  & \CheckmarkBold &  \\
E-mail &  &  &  &  &  &  & \CheckmarkBold & \CheckmarkBold \\
Letters &  &  &  &  &  &  & \CheckmarkBold & \CheckmarkBold \\
\hline
\end{tabular}%
}
\caption{Constraints of grounding along with their medium of communication \cite{DBLP:books/others/91/ClarkB91}}
\label{tab:constraints-possibilities}
\end{table*}

\paragraph{Purviews - Personalization: }
Upon a continous exchange regarding this topic (and perhaps a few other times earlier), the agent needs to adapt and personalize to the interacting human over time. 

\vspace{1em}
\hrule
\vspace{0.5em}
\noindent \textit{\textbf{Human}: Can you give me any movie suggestions?}

\noindent \textit{\textbf{Agent}: Yes, since you like Disney movies and seem interested in Robert Downey Jr,  would you like to watch ``Dolittle''?}
\vspace{0.5em}
\hrule
\vspace{1em}

Having discussed about Robert Downey Jr in prior contexts and retaining from the prior interactions that the human likes Disney movies, when the human asks about a movie recommendation, the agent continually learns and contextually suggests Robert Downey Jr's Disney movie ``Dolittle'' as a recommendation.

\paragraph{Constraints - Copresence: }
Modality is an important medium that affects communicative goals and the nature of interaction. Here is an example in a copresent environment. 

\vspace{1em}
\hrule
\vspace{0.5em}
\noindent \textit{\textbf{Human}: I want to play with my cat. Can you get me the ball on your right?}

\vspace{0.5em}
\hrule
\vspace{1em}

In the above example, the human and the agent are copresent in the same environment. The above utterance for instance, includes executable actions in the environment along with references being either person-centric or agent-centric.

\paragraph{Constraints - Visibility: }
Certain communications like in the cases of visual question answering or visual dialog only presents a visible medium to interact \textit{about}. The interaction requires information from an image or a video, but does not necessarily include executable actions or cater to external knowledge of the information. For example, with an access to an image a human can ask a question like the following:

\vspace{1em}
\hrule
\vspace{0.5em}
\noindent \textit{\textbf{Human}: How many peaks are there in those mountain ranges?}

\vspace{0.5em}
\hrule
\vspace{1em}

\paragraph{Constraints - Audibility: }
This modality constrains the information scope to be within speech signals that are only heard and do not contain any visual or copresent information. 

Table \ref{tab:constraints-possibilities} presents the constrainst of grounding.

\section{Further survey and categories}

Here is a brief elaboration of the datasets presented in Table \ref{tab:new-datasets}.

\paragraph{New datasets:} The first solution to curate the entire dataset with annotations designed for the task.

\noindent \textit{ $\bigcdot$ Non-textual Modality: }
For images, new datasets are curated for a variety of tasks including caption relevance \cite{DBLP:conf/acl/SuhrZZZBA19}, multimodal MT \cite{DBLP:conf/emnlp/ZhouCLY18},
soccer commentaries \cite{DBLP:conf/emnlp/Koncel-KedziorskiHF14}
semantic role labeling \cite{DBLP:conf/emnlp/SilbererP18}, instruction following \cite{DBLP:conf/eacl/SchlangenH17a}, 
navigation \cite{DBLP:conf/conll/AndreasK14}, understanding physical causality of actions \cite{DBLP:conf/acl/GaoDYC16}, understanding topological spatial expressions \cite{DBLP:conf/acl/KelleherKC06}, spoken image captioning \cite{DBLP:conf/conll/AlishahiBC17}, 
entailment \cite{DBLP:conf/coling/Vu0EJLTTBG18}, image search \cite{DBLP:conf/acl/KirosCH18}, scene generation \cite{DBLP:conf/acl/ChangMSPM15}, etc.,
Coming to videos, datasets have become popular for several tasks like identifying action segments \cite{DBLP:journals/tacl/RegneriRWTSP13}, sematic parsing \cite{DBLP:conf/emnlp/RossBBMK18}, instruction following from visual demonstration \cite{DBLP:conf/emnlp/LiuYSSHZC16}, spatio-temporal question answering \cite{DBLP:conf/acl/LeiYBB20}, etc.,

\begin{figure*}[tbh]
\centering
\includegraphics[trim=0cm 0cm 0cm 0cm, clip,width=0.92\linewidth]{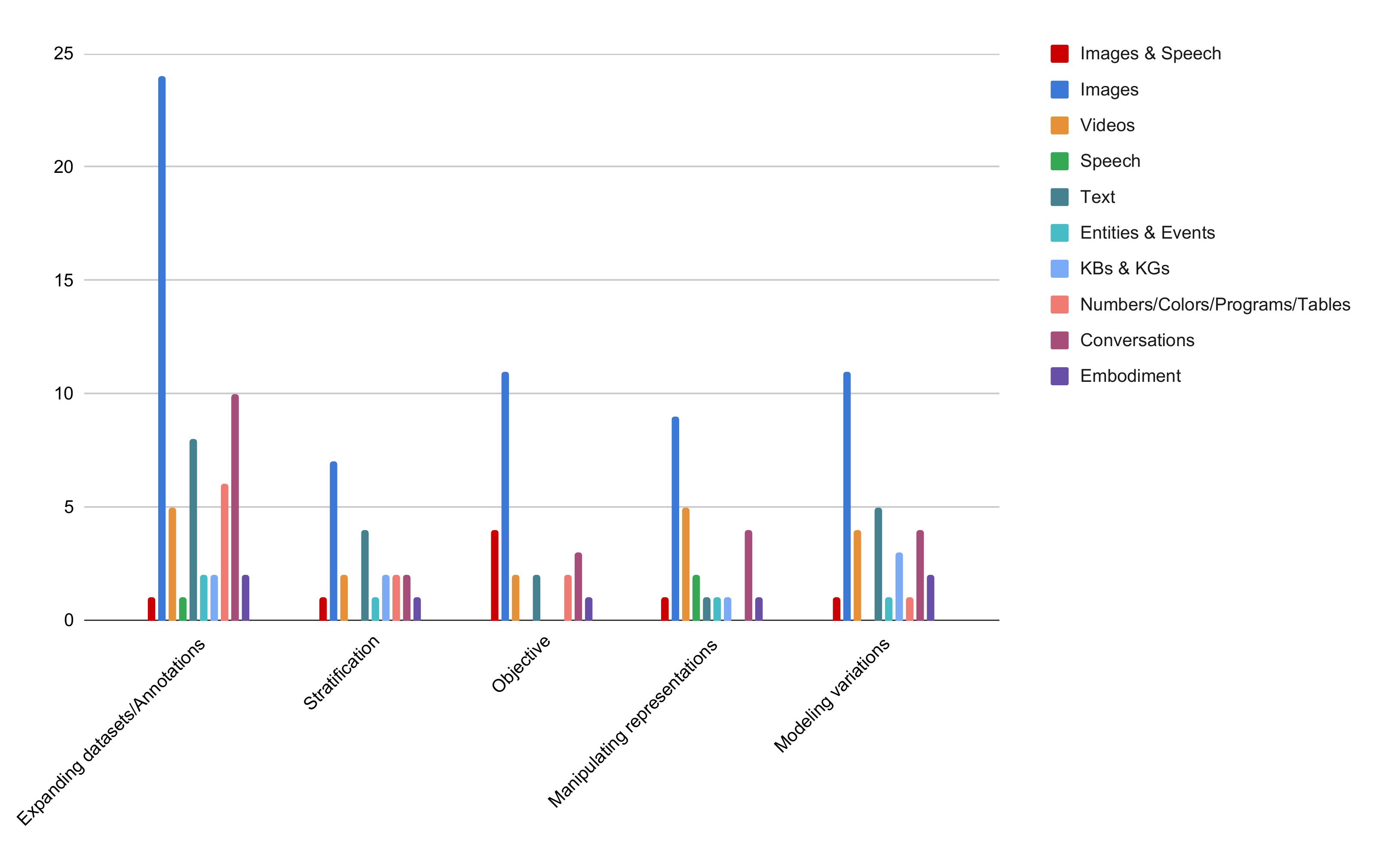}
\caption{Comprehensive trends of papers using differnt techniques in various modalities to address grounding}
\label{fig:app_modalities}
\end{figure*}

\noindent \textit{ $\bigcdot$ Textual Modality: }
Within text, there are several datasets for tasks like 
content transfer \cite{DBLP:conf/naacl/PrabhumoyeQG19}, 
commonsense inference \cite{DBLP:conf/emnlp/ZellersBSC18}, 
reference resolution \cite{DBLP:conf/acl/KenningtonS15}, 
symbol grounding \cite{DBLP:conf/emnlp/KamekoMT15}, 
studying linguistic and non-linguistic contexts in microblogs \cite{DBLP:conf/naacl/DoyleF15}, 
bilingual lexicon extraction \cite{DBLP:conf/coling/LawsMDSHS10}, 
universal part-of-speech tagging for low resource languages \cite{DBLP:conf/naacl/CardenasLJM19}, 
entity linking and reference \cite{DBLP:conf/acl/NothmanHHC12} etc.,

\noindent \textit{ $\bigcdot$ Other: }
More static grounding datasets correspond to tasks like  
identifying phrases representing variables \cite{DBLP:conf/emnlp/RoyUR16}, 
conceptual similarity in olfactory data \cite{DBLP:conf/acl/KielaBC15}, 
identifying colors from descriptions \cite{DBLP:journals/tacl/MonroeHGP17}, 
correcting numbers \cite{DBLP:conf/emnlp/SpithourakisAR16} etc.,

\noindent \textit{ $\bigcdot$ Interactive: }
Coming to an interactive setting, the datasets span tasks like conversations based on negotiations \cite{DBLP:conf/emnlp/CadilhacABL13}, 
referring expressions from images \cite{DBLP:conf/acl/HaberBTGBF19, takmaz2020refer},
emotions and styles \cite{DBLP:conf/acl/ShusterHBW20}, 
media interviews \cite{DBLP:conf/emnlp/MajumderLNM20}, documents \cite{DBLP:conf/emnlp/ZhouPB18}, improvisation \cite{DBLP:conf/acl/ChoM20}, problem solving \cite{DBLP:conf/naacl/LiB15}, 
spatial reasoning in a simulated environment \cite{DBLP:journals/tacl/JannerNB18}, 
navigation \cite{DBLP:conf/emnlp/KuAPIB20} etc.,

In addition, there are several other techniques used to ground phenomenon in real world contexts.

In addition to the techniques dicscussed in the paper, we also studied the categorization based on stratification, which is explained here.
\paragraph{Stratification: }
The stratification technique characterizes the input or the model to explicitly cater to the compositionality property. This can be done by either breaking down the input to meaningful compositions or building the model to compose the representations. Utilizing grammatical rules need not necessarily lead to compositions, although there is an overlap between these two techniques.


A common strategy when language is involved is leveraging \textbf{syntax and parsing}.
In the domain of images, 
\citet{DBLP:conf/emnlp/UdagawaYA20} 
design an annotation protocol to capture important linguistic structures based on predicate-argument structure, modification and ellipsis to utilize linguistic structures based on spatial expressions.
\citet{becerra2018gold} study linguistic complexity from a developmental point of view by using syntactic rules to provide data to a learner, that identifies the underlying language from this data.
\citet{DBLP:conf/acl/ShiMGL19} use image-caption pairs to extract constituents from text, based on the assumption that similar spans should be matched to similar visual objects and these concrete spans form constituents.
\citet{DBLP:conf/acl/KelleherKC06} use combinatory categorial grammar (CCG) to build a psycholinguistic based model to predict absolute proximity ratings to identify spatial proximity between objects in a natural scene.
\citet{DBLP:conf/emnlp/RossBBMK18} employ CCG-based parsing to a fixed set of unary and binary derivation rules to generate semantic parses for videos.

\noindent \textit{ $\bigcdot$ Textual Modality: }
\citet{DBLP:conf/acl/JohnsonDF12} study the modeling the task of inferring the referred objects using social cues and grammatical reduction strategies in language acquisition.
\citet{DBLP:conf/acl/Eckle-Kohler16} attempt to understand meaning in syntax by a multi-perspective semantic characterization of the inferred classes in multiple lexicons. \citet{DBLP:conf/acl/Chen12} develop a context-free grammar to understand formal navigation instructions that correspond better with words or phrases in natural language.
\citet{DBLP:conf/emnlp/BorschingerJJ11} study the probabilistic context-free grammar learning task using the inside-out algorithm in game commentaries.
CCG parsers are also used to perform entity slot filling task \cite{DBLP:conf/emnlp/BiskRBHS16}.
When applied to question answering over a database, dependency rules are used to model the edge states as well as transitions such as the work done by using a treeHMM \cite{DBLP:conf/acl/Poon13}.

\noindent \textit{ $\bigcdot$ Other: }
\citet{DBLP:conf/emnlp/RoyUR16} perform equation parsing that identifies noun phrases in a given sentence representing variables using high precision mathematical lexicon 
to generate the correct relations in the equations.
\citet{DBLP:conf/naacl/ParikhPT15} perform prototype driven learning to learn a semantic parser in tables of nested events and unannotated text.

\noindent \textit{ $\bigcdot$ Interactive: }
\citet{DBLP:journals/tacl/LuongFJ13} 
use parsing and grammar induction to produce a parser capable of representing full discourses and dialogs.
\citet{DBLP:conf/acl/Steels04} study games and embodied agents by modeling a constructivist approach based on invention, abduction and induction to language development.

Another frequently used technique when language is involved is by leveraging the principle of \textbf{compositionality}. This implies that the meaning of a complex expression is determined by the meanings of its constituents and how they interact with one another.

\noindent \textit{ $\bigcdot$ Non-textual Modality: }
In the domain of images, \citet{DBLP:conf/acl/SuhrZZZBA19} present a new dataset 
to understand challenges in language grounding including compositionality, semantic diversity and visual reasoning. \citet{DBLP:conf/acl/ShiMGL19}, discussed earlier also use grammar rules to compose the inputs. \citet{DBLP:conf/emnlp/Koncel-KedziorskiHF14} leverage the compositional nature of language to understand professional soccer commentaries.
In the domain of videos, \citet{DBLP:conf/coling/NayakM12} study language acquisition by segmenting the world to obtain a meaning space and combining them to get a linguistic pattern.

\noindent \textit{ $\bigcdot$ Textual Modality: }
With ontologies, \citet{DBLP:conf/emnlp/PappasMS20} perform adaptive language modeling to other domains to get a fully compositional output embedding layer 
which is further grounded in information from a structured lexicon.

\noindent \textit{ $\bigcdot$ Interactive: }
\citet{roy2003conversational} work on grounding word meanings for robots by composing perceptual, procedural, and affordance representations.



\paragraph{Hierarchical modeling} is also applied to show effect of introducing phone, syllable, or word boundaries in spoken captions \cite{DBLP:conf/conll/HavardBC20} and with a compact bilinear pooling in visual question answering \cite{DBLP:conf/emnlp/FukuiPYRDR16}.


\noindent There is some work that presents a bayesian probabilistic formulation to learn referential grounding in dialog  \cite{DBLP:conf/acl/LiuSFC14}, user preferences \cite{DBLP:conf/emnlp/CadilhacABL13}, color descriptions \cite{DBLP:journals/tacl/McMahanS15, DBLP:conf/conll/AndreasK14}.

\noindent A huge chunk of work also focus on leveraging attention mechanism for grounding multimodal phenomenon in images \cite{DBLP:conf/emnlp/SrinivasanSME20, DBLP:conf/coling/ChuON18, DBLP:conf/acl/HuangHGQZ19, DBLP:conf/acl/FanWWH19, DBLP:conf/coling/Vu0EJLTTBG18, DBLP:conf/acl/KawakamiDB19, DBLP:journals/corr/abs-1909-09699}, videos \cite{DBLP:conf/acl/LeiYBB20, DBLP:conf/acl/ChenMLW19} and navigation of embodied agents \cite{DBLP:conf/emnlp/YangLN20}, etc.,

\noindent Some approach this using data structures such as graphs in the domains of grounding images \cite{DBLP:conf/acl/ChangMSPM15, DBLP:conf/acl/LiuSFC14}, videos \cite{DBLP:conf/emnlp/LiuYSSHZC16}, text \cite{DBLP:conf/coling/LawsMDSHS10, DBLP:conf/acl/Chen12, masse2008meaning}, entities \cite{DBLP:conf/emnlp/ZhouKTR18}, 
knowledge graphs and ontologies \cite{DBLP:conf/naacl/JauharDH15, DBLP:conf/acl/ZhangLXL20} and interactive settings \citet{DBLP:conf/naacl/JauharDH15, DBLP:conf/acl/XuWNWCL20}.

Here is the technique wise representation of these categories of models in the literature.

\begin{figure}[H]
\centering
\includegraphics[trim=0cm 0cm 0cm 0cm, clip,width=0.85\linewidth]{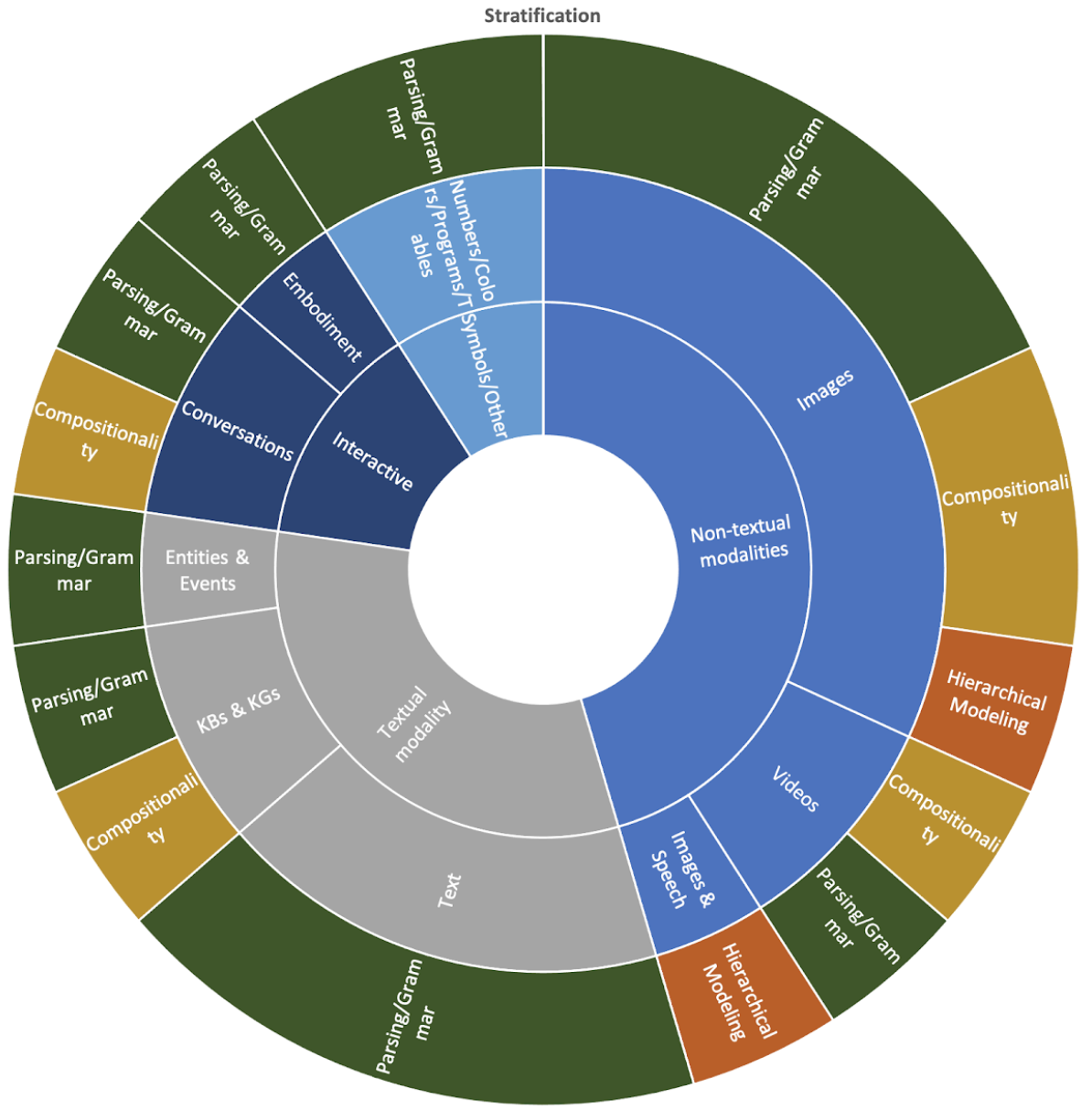}
\caption{Papers addressing stratification in grounding}
\label{fig:app_constraints}
\end{figure}

\section{Prevelance of modailties and constraints}

Here is the distribution of the papers studying various tasks based on the constraints imposed by the medium.
\begin{figure}[H]
\centering
\includegraphics[trim=0cm 0cm 0cm 0cm, clip,width=0.99\linewidth]{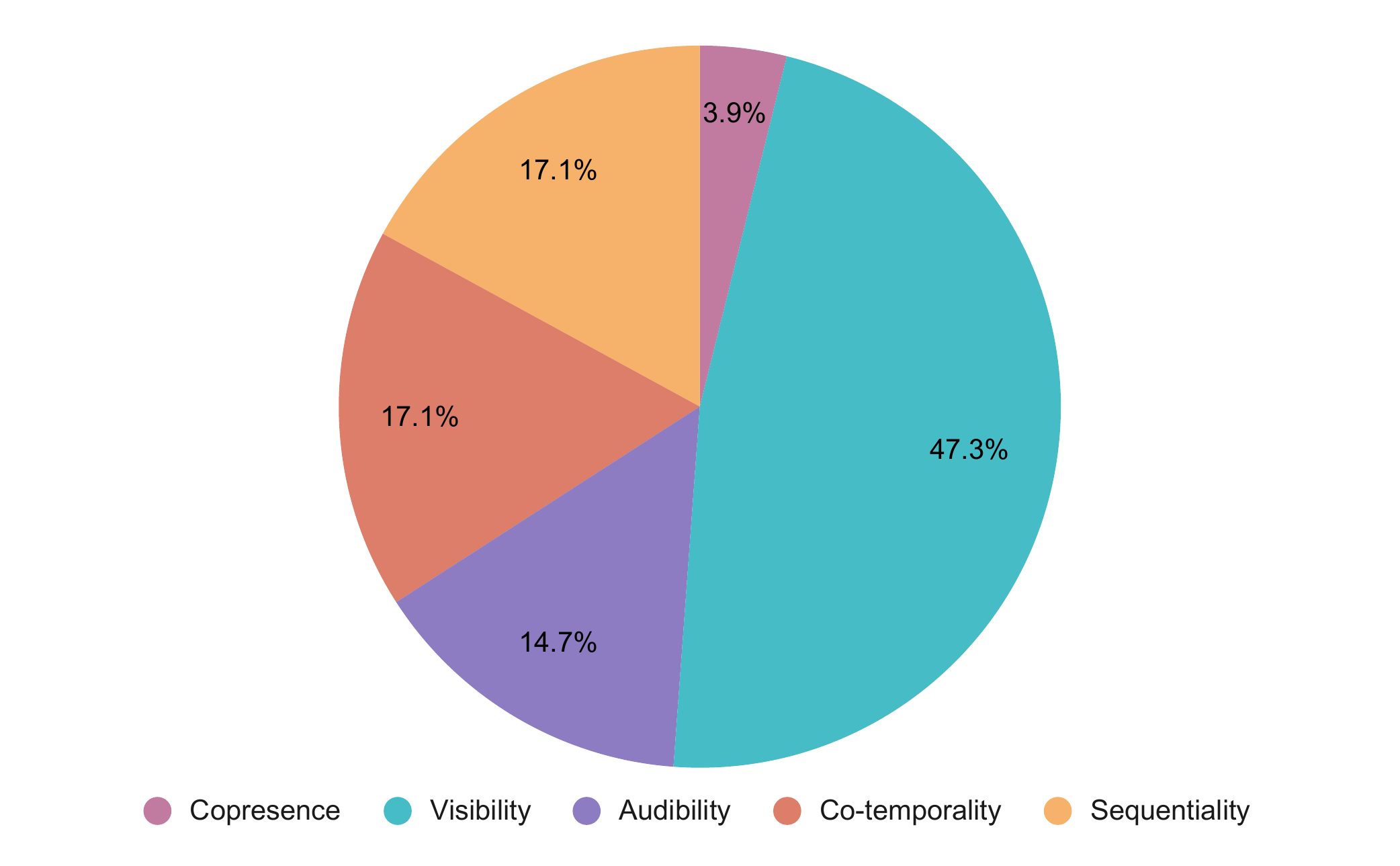}
\caption{Papers addressing different constraints of grounding}
\label{fig:app_constraints}
\end{figure}

As we can see, a major concentration of these efforts lie in grounding visual and textual media, while a few cater to audibility i.e speech signals. Papers studying dialog are the main representatives of the constraints for sequentiality and co-temporality. 

\section{Objective function for grounding}

\begin{figure}[H]
\centering
\includegraphics[trim=0cm 0cm 0cm 0cm, clip,width=0.85\linewidth]{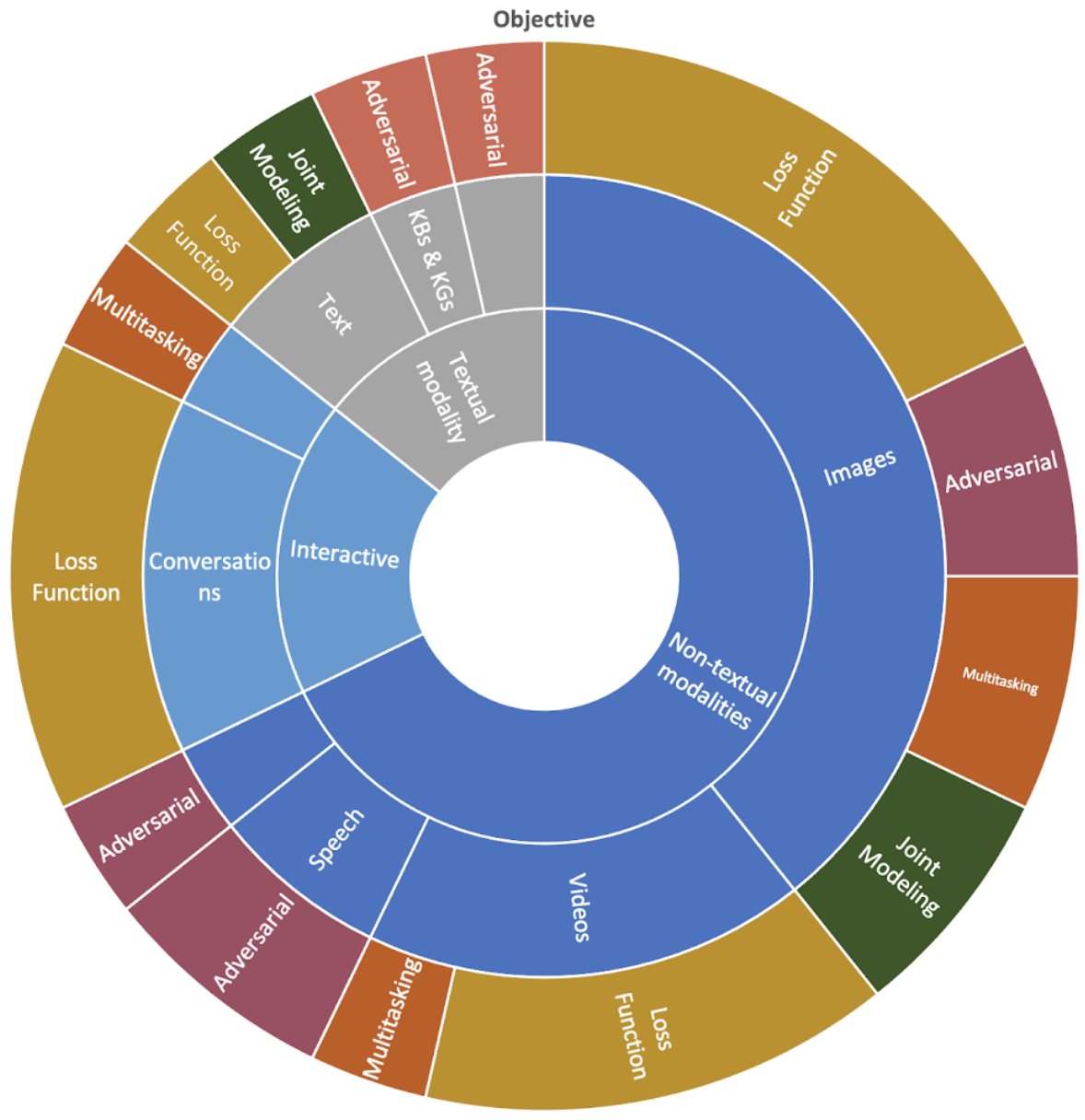}
\caption{Objective functions in papers studying grounding}
\label{fig:app_objective}
\end{figure}

Figure \ref{fig:app_objective} systeamtically presents the trends in the usage of the objective function for grounding as discussed in Section \ref{sec:approaches}. While manipulating loss function is a stable trend, recent approaches are focusing on adversarial objectives to perform grounding.

\section{Nuanced modeling variations for grounding}

\begin{figure}[H]
\centering
\includegraphics[trim=0cm 0cm 0cm 0cm, clip,width=0.85\linewidth]{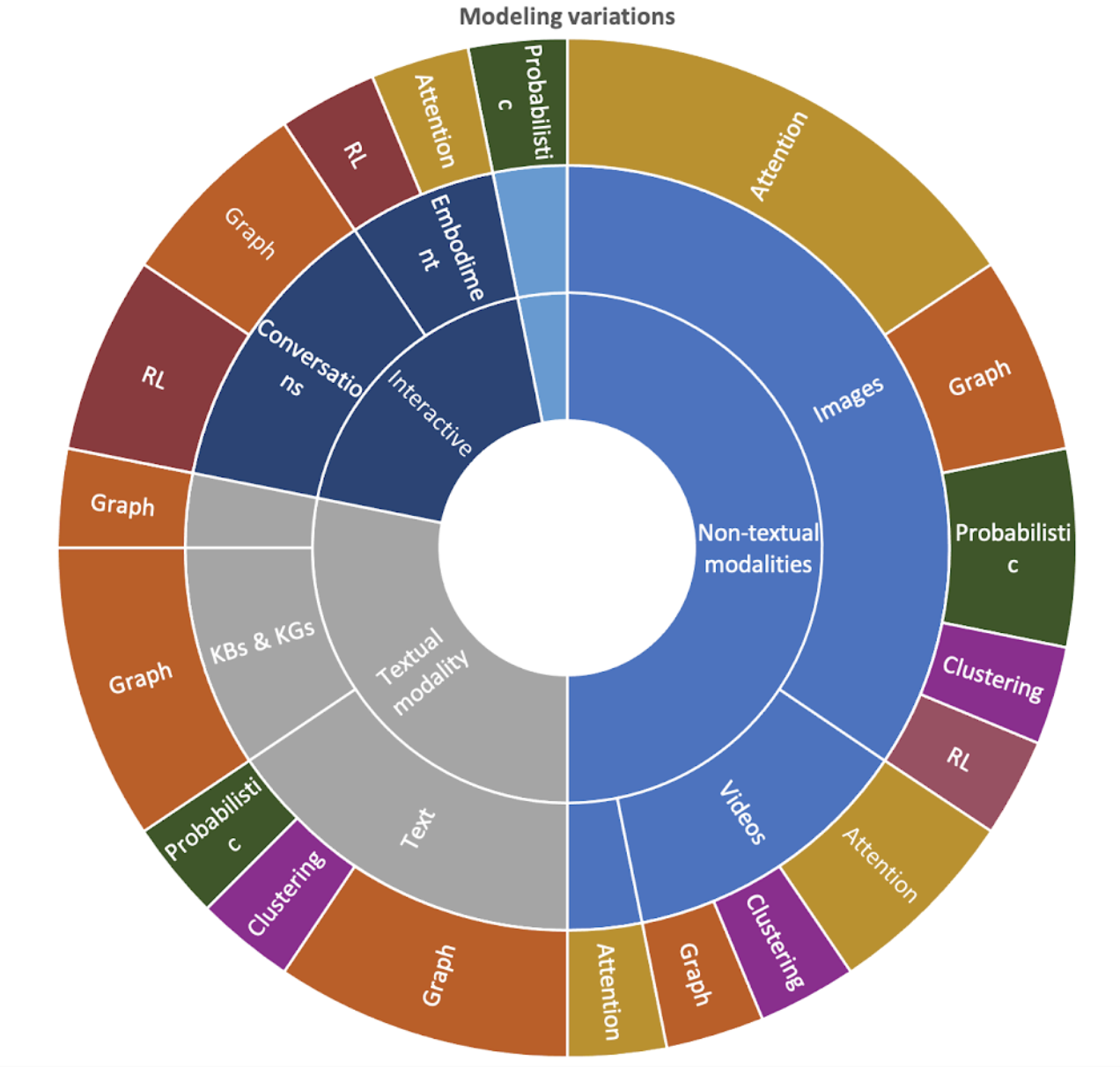}
\caption{Modeling variations in papers studying grounding}
\label{fig:app_modeling_variations}
\end{figure}

Here is a more nuanced and finer grained categorization of the various modeling techniques used in literature for grounding. Figure \ref{fig:app_modeling_variations} presents these categories in depth.

As discussed in the paper, most of the literature is focused on grounding in static visual modality. Attention based methods dominate the rest of the methods in both textual and non-textual modalities closely followed by graph based methods as observed in these trends.


This is not an exhaustive study of all the techniques that present grounding, but are some of the representative categories. Here are more studies that perform grounding with various techniques such as clustering \cite{DBLP:conf/acl/ShutovaTM15, DBLP:conf/naacl/CardenasLJM19} 
regularization \cite{DBLP:conf/acl/ShresthaKK20}, 
CRFs \cite{DBLP:conf/acl/GaoDYC16}, 
classification \cite{pangburn-etal-2003-ebla, DBLP:journals/tacl/MonroeHGP17}, 
linguistic theories \cite{DBLP:journals/coling/StrubeH99}, 
iterative refinement \cite{DBLP:conf/acl/LiNMFLZ19, DBLP:conf/interspeech/ChanduB20}, 
language modeling \cite{DBLP:conf/emnlp/SpithourakisAR16, DBLP:conf/acl/ChoM20}, 
nearest neighbors \cite{DBLP:conf/acl/KielaBC15}, contextual fusion \cite{chandu2019my},
mutual information \cite{oates2003grounding}, 
cycle consistency \cite{DBLP:conf/emnlp/ZhongLWZ20} etc.,

\end{document}